\documentclass[runningheads]{llncs}
\usepackage{graphicx}
\usepackage{xcolor}
\usepackage{amsmath}
\begin{document}
\title{Parameter-Linear Reformulation of the Time-Optimal Path Following Problem with Arbitrary Continuity}
\titlerunning{Parameter-Linear Time-Optimal Path Following}
%
\author{Tobias Marauli\orcidID{0000-0003-1117-0315} \and Hubert Gattringer\orcidID{0000-0002-8846-9051} \and Andreas M{\"u}ller\orcidID{0000-0001-5033-340X}}
\authorrunning{T. Marauli et al.} 
%
\institute{Institute of Robotics, Johannes Kepler University Linz, \\Altenberger Strasse 69
	4040 Linz, Austria\\
	\email{\{tobias.marauli, hubert.gattringer, a.mueller\}@jku.at}
} 
\maketitle              
\begin{abstract}
In this paper the challenges of time-optimal path following are addressed. A conventional approach is to minimize the travel time, which inevitable leads to singularities at zero path speed, when reformulating the optimization problem in terms of a path parameter. Thus, smooth trajectory generation while maintaining a low computational effort is quiet challenging, since the singularities have to be taken into account.  To this end, a different approach is presented in this paper. This approach is based on maximizing the path speed along a prescribed path. Furthermore, the approach is capable of planning smooth trajectories numerically efficient. Moreover, the discrete reformulation of the underlying problem is linear in optimization variables. 
	
\end{abstract}

\section{INTRODUCTION}

Short duration times are highly requested to maximize the productivity of a robotic system. Therefore, time optimal motion is desirable. Frequently, the task for a robot is defined by a path that should be followed. Examples are welding and painting robots, CNC-machines as well as autonomous vehicles. Thus, several approaches for following paths in a (time-) optimal way are made in the literature \cite{Kaserer2019,OberherberM,Pfeiffer1987,SlotineEffic,MinimumTimeKang}. A conventional approach is to reformulate the optimization problem, where the time is minimized, in terms of a path parameter. Solving the reformulated problem can be carried out numerically efficient as adopted in \cite{ConvOptAppr,TimeOptConvConc}. Using this approach comes with several drawbacks. First, singularities due to the reformulation in terms of the path parameter are present, which have to be considered. Consequently, the use of an arbitrary transcription method e.g. collocation \cite{PathConstrTraj} or shooting methods \cite{BOCK19841603,TrajPlannColl} while maintaining low computation times is quite challenging. To this end, a different approach to minimize the travel time is presented in this paper. This approach is based on maximizing the path speed along the path. The latter approach is free of singularities, which copes well with arbitrary transcription methods for smooth trajectory planning. Moreover, the resulting discrete optimization problem is linear in the approach of the path parameter.


\section{Optimal path following problem formulation}
\label{sec:NewApproach}

In this section the parameter-linear optimal path following (OPF) problem is derived. To this end, consider a $n$-DOF robotic manipulator with $\mathbf{q} \in S^{n}$ denoting the vector of joint coordinates. The equations of motion (EOM)
\begin{equation}
\mathbf{M}(\mathbf{q}) \ddot{\mathbf{q}} + \mathbf{h}(\mathbf{q}, \dot{\mathbf{q}}) = \mathbf{\tau}
\label{eq:EOM}
\end{equation}	
describe the dynamic behavior of the manipulator, where $\mathbf{M}(\mathbf{q})$ is the mass matrix and $ \boldsymbol{\tau}$ are the joint torques. The Coriolis, centrifugal and gravitational terms as well as the Coulomb and viscous friction are described by the vector $\mathbf{h}(\mathbf{q}, \dot{\mathbf{q}})$.  
If not necessary, the dependencies are omitted for simplicity and
readability. Without loss of generality, a geometric joint path $\mathbf{q}(\sigma)$ is supposed to be given and parameterized in terms of a path parameter ${\sigma \in \left[ 0,1\right]}$, with the geometric defined as $()^\prime = \partial()/\partial\sigma$.
The time dependency of the path follows from the relation $\sigma(t)$ between the path parameter and the time. Furthermore, it is assumed that the path is followed only in forward direction ${\dot{\sigma}(t) \geq 0}$, ${\forall t \in\left[ 0,t_\text{T}\right]}$. Moreover, the introduction of the squared path speed  ${z(\sigma)  := \dot{\sigma}^2}$ is beneficial for the optimization as adopted in \cite{ConvOptAppr,Debrouwere2013}. The introduction of latter leads to the relations
\begin{equation}
\dot{\sigma} = \sqrt{z}, \quad \ddot{\sigma} = \frac{z^\prime}{2}, \quad \dddot{\sigma} = \frac{z^{\prime \prime}}{2}\sqrt{z} 
\label{eq:Sigma_Z_relation}
\end{equation}
for higher time derivatives of $\sigma$. Using the chain rule and upper relations the joint velocities, accelerations and jerks along a prescribed path ${\mathbf{q}(\sigma)}$ can be written as 
\par\vspace{-\baselineskip}
\begin{subequations}
	\label{eq:JerkAccVel}	
	\begin{align}
	&\dot{\mathbf{q}}(\sigma)  = \mathbf{q}^\prime(\sigma)  \sqrt{z} \label{eq:JointVel}\\
	&\ddot{\mathbf{q}}(\sigma)  = \mathbf{q}^{\prime \prime}(\sigma)  z +  \frac{1}{2}\mathbf{q}^\prime(\sigma)   z^\prime \label{eq:JointAcc}\\
	&\dddot{\mathbf{q}}(\sigma)  = (\mathbf{q}^{\prime \prime \prime}(\sigma)  z + \frac{3}{2}\mathbf{q}^{\prime \prime}(\sigma)  z^\prime  + \frac{1}{2}\mathbf{q}^\prime(\sigma)   z^{\prime \prime}) \sqrt{z}. \label{eq:JointJerk}	
	\end{align}	
\end{subequations}
The substitution of \eqref{eq:JointVel} and \eqref{eq:JointAcc} in \eqref{eq:EOM} yields the path projected EOM \cite{ConvOptAppr}
\begin{equation}
\mathbf{a}(\sigma) z^\prime + \mathbf{b}(\sigma) z + \mathbf{c}(\sigma) + \mathbf{d}(\sigma)\sqrt{z} =  \boldsymbol{\tau}(\sigma), \label{eq:TrqPath}
\end{equation}
where $ \mathbf{a}(\sigma)$ is related to the mass matrix and $\mathbf{b}(\sigma)$ includes the path projected Coriolis and centrifugal terms,  mixed with higher path derivative terms. The path depending gravitational and the Coulomb friction terms are described by $\mathbf{c}(\sigma)$ and $\mathbf{d}(\sigma)$ denotes the path projected viscous friction.

For planning time-optimal trajectories the conventional approach of minimizing the terminal time
\begin{equation}
 t_\text{T} = \int_{0}^{t_\text{T}} 1\,\text{d}t = \int_{0}^{1} \frac{1}{\sqrt{z(\sigma)}}\, \text{d}\sigma, \label{eq:TimeEqu}
\end{equation}
is primary used \cite{TimeEnergShiller,MinimumTimeKang,SlotineEffic}. 
Due to the division by ${\sqrt{z(\sigma)}}$ in above integral singularities, where ${z(\sigma) = 0}$, have to be taken into account. This has to be considered when using numerical methods e.g. shooting or collocation methods for discretizing the time-optimal path following problem. Therefore, time-optimal trajectory planning with arbitrary smoothness while maintaining low computational effort, is quiet challenging. For example, the numerically efficient approach presented in \cite{ConvOptAppr,Debrouwere2013} where the time-optimal path following problem is reformulated as second-order-cone problem is not capable of planning trajectories with arbitrary smoothness, due to the aforementioned singularities.

Planning time-optimal trajectories inevitable leads to increasing the speed $z$ along the path, to minimize the required terminal time \eqref{eq:TimeEqu} as adopted in \cite{Kaserer2019}. An upper bound of $z$ is given by the maximum velocity curve (MVC) \cite{Pfeiffer1987}, defined by the velocity, acceleration and torque constraints. Therefore, maximizing the path speed $z$ gives rise to a different approach of reducing the terminal time $t_\text{T}$. Since ${z \geq 0}$, the integral of the path speed
\begin{equation}
\int_{0}^{1} z(\sigma) \, \text{d}\sigma, \label{eq:ZIntegral}
\end{equation}
is a concave functional regarding ${z(\sigma)}$ \cite{rockafellar1968integrals}. Using above integral no singularities have to be taken into account, since the path speed is bounded and no division is present. The optimization problem of latter approach can be written as
\begin{align}
&\displaystyle \max_{z(\sigma)}   \,  \int_{0}^{1} z \, \text{d} \sigma,   \nonumber \\
&\text{s.t.} ~~
 \underline{\dot{\mathbf{q}} } \leq \dot{\mathbf{q}}(\sigma)  \leq \overline{\dot{\mathbf{q}} }, ~~ \quad
\underline{\ddot{\mathbf{q}} } \leq \ddot{\mathbf{q}}(\sigma)  \leq \overline{\ddot{\mathbf{q}} }, \nonumber \\
&\hphantom{s.t.}~\underline{\dddot{\mathbf{q}} } \leq \dddot{\mathbf{q}}(\sigma) \leq \overline{\dddot{\mathbf{q}} }, ~~
\underline{ \boldsymbol{\tau} } \leq  \boldsymbol{\tau}(\sigma)  \leq \overline{ \boldsymbol{\tau} }, \nonumber \\
&\hphantom{s.t.}~ z(0) = \dot{\sigma}_0^2, \qquad \quad ~~ z(1) = \dot{\sigma}_\text{T}^2,   \nonumber \\
&\hphantom{s.t.}~z(\sigma) \geq 0, \qquad \qquad \text{for} ~\sigma \in [0,1], \label{eq:MaxZ_OPF}
\end{align}
with the joint velocity, acceleration and jerk \eqref{eq:JerkAccVel} and the generalized torques \eqref{eq:TrqPath} as constraints. The lower and upper bounds of the constraints are denoted by $\underline{()},~\overline{()}$. Solving the problem yields a maximized squared path speed profile ${z(\sigma)^*}$. Since, optimization problems are solved numerically Sec. \ref{sec:ParamLinRefor} covers the discrete reformulation of \eqref{eq:MaxZ_OPF}.


\section{Parameter-linear reformulation}
\label{sec:ParamLinRefor}
This section shows that the discrete reformulation of \eqref{eq:MaxZ_OPF} leads to a parameter linear optimization problem, when the joint jerk constraints \eqref{eq:JointJerk} and the viscous friction in \eqref{eq:TrqPath} i.e. ${\mathbf{d}(\sigma) = \mathbf{0}}$ are not considered. To this end, first, the path parameter is discretized on ${[0,1]}$, with $N+1$ grid points ${\sigma_k \in \left[0,1\right]}$, for ${k= 0\dots N}$. Secondly, the function ${z(\sigma)}$ is modeled by a finite number of variables. To point out that the parameter linear problem formulation is received with an arbitrary approach for the path speed 1) a polynomial and 2) a B-spline approach are considered in this paper. 

\subsection{Piecewise-polynomial approach}

The piecewise-polynomial approach is capable of planning optimal trajectories with arbitrary smoothness, by choosing the degree of the polynomial. For example a polynomial of degree three, leads to smooth joint jerk trajectories. In order to validate the presented approach in terms of the computation time and optimization result quality, the same approach as used in \cite{ConvOptAppr,TimeEnVersch}, for discretizing the conventional time-optimal problem, is used. The piecewise-linear approach is given by
\begin{equation}
z(\sigma) = z_k + (z_{k+1}- z_k)\frac{\sigma - \sigma_k}{\sigma_{k+1} - \sigma_k}
\label{eq:zLinAnsatz}
\end{equation}
for ${\sigma \in  \left[\sigma_k,\sigma_{k+1}\right]}$, where the optimization variables are assigned on the grid points ${z(\sigma_k) = z_k}$. Similar as in \cite{TimeEnVersch}, the function $z^\prime(\sigma)$ is evaluated in between the grid points, namely ${\sigma_{k+1/2} = (\sigma_{k+1} + \sigma_k)/2}$. Thus, the joint acceleration \eqref{eq:JointAcc} and torque constraints \eqref{eq:TrqPath} are evaluated at $\sigma_{k+1/2}$ accordingly.
By introducing  \eqref{eq:zLinAnsatz} in  \eqref{eq:JointVel} the joint constraints can be reformulated as linear constraints, by carefully squaring them similar to \cite{ConvOptAppr}. Substituting the piecewise-linear approach \eqref{eq:zLinAnsatz} in  \eqref{eq:JerkAccVel}, \eqref{eq:TrqPath} and \eqref{eq:ZIntegral} the discrete optimization problem is written as
\begin{align}
&\displaystyle \max_{z_k}   \,  \sum_{k=0}^{N-1} \frac{1}{2}(z_k + z_{k+1})\Delta \sigma_k,   \nonumber \\
&\text{s.t.} ~~
\mathbf{q}^\prime(\sigma_k)^2 z_k \leq \overline{\dot{\mathbf{q}}}^2 ~ \text{for} ~\mathbf{q}^\prime(\sigma_k) \geq 0, ~  \mathbf{q}^\prime(\sigma_k)^2 z_k \leq \underline{\dot{\mathbf{q}}}^2 ~ \text{for} ~\mathbf{q}^\prime(\sigma_k) < 0, \nonumber \\
&\hphantom{s.t.}~~ \underline{\ddot{\mathbf{q}} } \leq \ddot{\mathbf{q}}(\sigma_{k+1/2}) \leq \overline{\ddot{\mathbf{q}} }, ~ \underline{\dddot{\mathbf{q}} } \leq \dddot{\mathbf{q}}(\sigma_{k+1/2}) \leq \overline{\dddot{\mathbf{q}} },~\underline{ \boldsymbol{\tau} } \leq  \boldsymbol{\tau}(\sigma_{k+1/2})  \leq \overline{ \boldsymbol{\tau} }, \nonumber \\
&\hphantom{s.t.}~~ z_0 = \dot{\sigma}_0^2, ~ z_N = \dot{\sigma}_\text{T}^2, ~z_k \geq 0, ~ \text{for} ~k = 0\dots N, \label{eq:MaxZ_OPF1}
\end{align}
with ${\Delta \sigma_k = \sigma_{k+1} -  \sigma_k}$. In order to increase the readability only, the joint velocity constraints are shown exemplary. The latter optimization problem is linear in $z_k$, without considering the joint jerk constraints and the viscous friction in \eqref{eq:TrqPath} i.e. ${\mathbf{d}(\sigma) = \mathbf{0}}$. Since linear problems can be solved numerically efficient, a sequential convex programming (SCP) \cite{Debrouwere2013,TimeOptConvConc} framework is used to incorporate the joint jerk constraints and the viscous friction in the torque constraints in the optimization process.

\subsection{B-spline approach} 
A more flexible approach of representing the path speed results from the use of B-splines, since the number of  control points can be varied independently from the number of grid points $N$ and the degree of the basis functions. Furthermore, B-splines were successfully used in smooth trajectory planning as discussed in \cite{GASPARETTO2007455}. Thus, the path speed represented by B-splines is written as
\begin{equation}
z(\sigma) = \sum_{i=0}^{n} p_i B^d_i(\sigma),
\label{eq:zBSplines}
\end{equation}
where $p_i$ are the control points, $n$ number of control points and ${B^d_i(\sigma)}$ are the basis functions of degree $d$. Introducing \eqref{eq:zBSplines} in \eqref{eq:JerkAccVel}, \eqref{eq:TrqPath} and \eqref{eq:ZIntegral} the discrete optimization problem yields to
\begin{align}
&\displaystyle \max_{p_i}   \,  \sum_{i=0}^{n} p_i \int_{0}^{1}\,  B^d_i(\sigma) \,  \text{d}\, \sigma,   \nonumber \\
&\text{s.t.} ~~
\underline{\dot{\mathbf{q}} } \leq \dot{\mathbf{q}}(\sigma)  \leq \overline{\dot{\mathbf{q}} }, ~ \underline{\ddot{\mathbf{q}} } \leq \ddot{\mathbf{q}}(\sigma_{k}) \leq \overline{\ddot{\mathbf{q}} }, ~ \underline{\dddot{\mathbf{q}} } \leq \dddot{\mathbf{q}}(\sigma_{k}) \leq \overline{\dddot{\mathbf{q}} }, \nonumber \\
&\hphantom{s.t.}~~\underline{ \boldsymbol{\tau} } \leq  \boldsymbol{\tau}(\sigma_{k})  \leq \overline{ \boldsymbol{\tau} }, ~ z_0 = \dot{\sigma}_0^2, ~ z_N = \dot{\sigma}_\text{T}^2,  ~z_k \geq 0, ~ \text{for} ~k = 0\dots N, \label{eq:MaxZ_OPF2}
\end{align}
with a linear objective function in the control points $p_i$. The joint velocity and acceleration constraints are also linear in $p_i$. Thus, problem \eqref{eq:MaxZ_OPF2} corresponds to a linear optimization problem without considering the joint jerk constraints and the viscous friction in \eqref{eq:TrqPath} i.e. ${\mathbf{d}(\sigma) = \mathbf{0}}$. To incorporate the latter, again a SCP approach is purposed.

\section{NUMERICAL EXAMPLE}

\subsection{Smooth trajectory planning}
\label{sec:SmoothTraj}
We used an Intel(R) Core(TM) i5-9500 CPU @ 3.00GHz CPU running on Windows 10 for all computations. For a 6-DOF \emph{Comau Racer3} robotic manipulator the optimal path following problem \eqref{eq:MaxZ_OPF} is solved in the parameter-linear form within a SCP framework for considering the joint jerk and torque constraints. The latter was implemented in \textsc{Matlab} using YALMIP \cite{Loefberg2004}. As solver MOSEK \cite{ApS2022}, a tailored solver for linear, quadratic, conic  problems is chosen. For simplicity the upper and lower bounds of \eqref{eq:MaxZ_OPF} are chosen symmetrically i.e. $ \underline{()} = -\overline{()}$.

\begin{figure}[htb!]
	\centering
	\includegraphics[trim = 1cm 0.4cm 1cm 0]{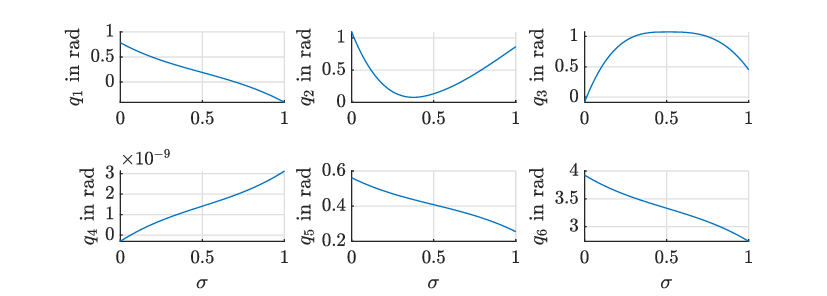}	
	\caption{Arbitrary path in the joint space ${\mathbf{q}(\sigma)}$.}
	\label{fig:JointPath}
\end{figure}

\begin{figure}[htb]	
	\begin{minipage}[b]{0.4\linewidth}
		\centering
		\includegraphics[trim = 0.5cm 0 0.7cm 0]{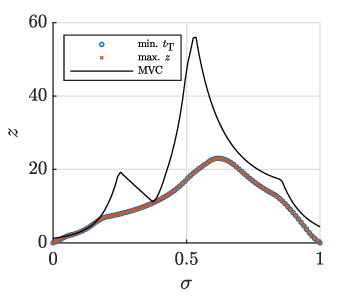}
		
	\end{minipage}
	\begin{minipage}[b]{0.6\linewidth}
		\centering
		\includegraphics[trim = 0.5cm 0 1cm 0]{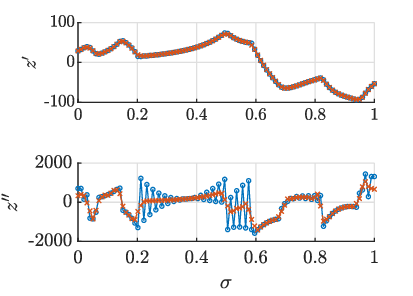}
	\end{minipage}
	\caption{Optimal path speed $z^*$ as well as first and second geometric derivative for following the arbitrary joint path. Blue circles: optimal solution with \eqref{eq:MaxZ_OPF1} using a piecewise-linear approach in $z$. Orange crosses: optimal solution with \eqref{eq:MaxZ_OPF2} using B-splines.}
	\label{fig:MVC_z_zst}
\end{figure}

As example for smooth optimal trajectory planning while retaining small computation times the joint path as illustrated in Fig. \ref{fig:JointPath} is picked. The latter path is designed with smooth derivatives $\mathbf{q}^{\prime},\, \mathbf{q}^{\prime \prime}$ and $\mathbf{q}^{\prime \prime \prime}$ for smooth jerk trajectory generation. Thus, $z,\, z^\prime,\, z^{\prime \prime}$ and $z^{\prime \prime \prime}$ have to be smooth for joint jerk trajectories, see \eqref{eq:JerkAccVel}. To this end, the OPF problem is solved with  1) the linear approach in $z$ \eqref{eq:MaxZ_OPF1} and 2) the B-spline ($d= 3$, $n=61$)  approach \eqref{eq:MaxZ_OPF2} with $N = 100$. Obviously the piecewise-linear approach in $z$ is not capable of planning smooth jerk trajectories, which is why it is only used as baseline strategy for the increase of the computational effort. The optimal solution $z^*$ of the two OPF problems is depicted in Fig. \ref{fig:MVC_z_zst}. Viewing the evolution of $z^*(\sigma)$ and the first derivative it is evident that both solutions are nearly the same. The resulting second derivative is only smooth with the B-spline approach \eqref{eq:MaxZ_OPF2} and thus results in smooth joint jerk trajectories, since ${\mathbf{q}(\sigma)}$ is smooth up to the third derivative. The resulting terminal and computation times with 1) are $t_\text{T} =0.429\,\text{s} $ and $t_\text{comp.} = 0.515\, \text{s}$ and 2) are $t_\text{T} =0.427\,\text{s} $ and $t_\text{comp.} = 0.899\, \text{s}$. As the results show, the computational effort is not increasing drastically as for the conventional approach, which is well discussed in \cite{ConvOptAppr,Kaserer2019,SmoothTraj}. Since the path speed is maximized up to the MVC while satisfying the considered constraints, at least a sub time-optimal solution is achieved with the presented approach \eqref{eq:MaxZ_OPF}.

\subsection{Presented vs Conventional approach}

In this section the 1) conventional approach ($\displaystyle \min t_\text{T}$), adopted in \cite{ConvOptAppr,TimeEnVersch}, is compared with the 2) presented approach \eqref{eq:MaxZ_OPF1} (using the piecewise-linear approach in $z$) in terms of computation and terminal time. The corresponding time
values are depicted in Tab. \ref{tab:StraightLineOptTab}, for different number of samples $N$ and four paths. The viewed paths include A) the arbitrary joint path, B) a rectangle with rounded corners (inspired by the ISO-9283 norm) as EE-path, C) a meander shaped EE-path and D) an arbitrary spatial curve. Due to the lack of space, the reader is referred to \cite{RobJournal_Marauli} for more information of the last three paths. In addition, the relative difference $p_{t_\text{T}}$ and ${t_\text{comp.}}$. compared to the conventional approach are also listed in the table. Thus, negative numbers reflect a loss in time (e.g. higher values of $t_\text{T}$) and vice versa. Concluding the results in Tab. \ref{tab:StraightLineOptTab} , the terminal of all four viewed paths is equal for the 1) conventional and 2) presented approach. Further, the resulting trajectories are also equal, which is not be shown due to the lack of space. These four paths show that \eqref{eq:MaxZ_OPF} is capable of computing time optimal solutions. To the best authors' knowledge, there exists no proof that the two optimization problems are equal. Nevertheless, the computational effort is reduced by more than 50\% when using \eqref{eq:MaxZ_OPF1}, which is a huge improvement in computing (sub) time-optimal trajectories. This also holds for planning trajectories with arbitrary smoothness as indicated in Sec. \ref{sec:SmoothTraj}.
\begin{table}[htb]
	\caption{Numerical values of the terminal and computation time as well as their relative differences compared to the conventional time-optimal approach.}
	\label{tab:StraightLineOptTab}
	\begin{center}
		\begin{tabular}{@{\extracolsep{\fill}}|l||ll||l|l||l|l||l|l||l|l|}
			\hline
			
			& & & \multicolumn{2}{l||}{A)} &  \multicolumn{2}{l||}{B)} & \multicolumn{2}{l||}{C)} & \multicolumn{2}{l|}{D)} \\  \cline{4-11}						
			& & & 1)      & 2)  &   1)      & 2) & 1)      & 2)  &   1)      & 2) \\ \hline
			
			$N=\,$100  &$t_\text{T}$ & in $ \text{s}$ & 0.429 & 0.429 &0.858 &0.858 & 1.936 &1.936 &0.892 &0.892  \\ \cline{2-11}
			& $t_\text{comp.}$ & in $ \text{s}$ & 4.969 & 0.490& 7.363 & 1.030 &8.733 &1.781 &5.047 &0.650  \\ \cline{2-11}
			& $p_{ t_\text{T}}$ & in $\%$  &- & 0.000 &- & 0.000&- &0.000 &- & 0.000  \\\cline{2-11}
			& $p_{t_\text{comp.}}$ & in $\%$ &- & 90.132 & -&86.014 &- &79.602 &- &87.120 \\ \hline
			
			$N=\,$200  &$t_\text{T}$ & in $ \text{s}$  &0.428 &0.428 & 0.860 &0.860 &1.958 &1.958 &0.893 &0.893 \\ \cline{2-11}
			& $t_\text{comp.}$ & in $ \text{s}$ &7.019 &0.623 & 14.089& 1.502&11.663 &2.647 &11.248 &1.251 \\ \cline{2-11}
			& $p_{ t_\text{T}}$ & in $\%$  &- & 0.000 & -& 0.000&- &0.000 &- & 0.000 \\\cline{2-11}
			& $p_{t_\text{comp.}}$ & in $\%$ &- & 91.117 &- &89.34 &- &77.303 &- &88.877 \\ 
			
			\hline
		\end{tabular}
	\end{center}
\end{table}

%
%
%

\section{CONCLUSION AND OUTLOOK}

This paper introduces a different approach of minimizing the terminal time $t_\text{T}$ by maximizing the speed $z$ along the path. In the presented method no singularities due to zero speed $z = 0$, have to be considered. Furthermore, it is shown that an arbitrary approach in the path speed (for smooth trajectory planning) results in a parameter linear problem reformulation, when the joint jerk constraints and viscous friction in \eqref{eq:TrqPath} i.e. ${\mathbf{d}(\sigma) = \mathbf{0}}$ are not considered. The comparison of the computation time in Tab.\ref{tab:StraightLineOptTab} between the presented and the conventional approach shows a overall reduction of more than $50\%$, while keeping the same terminal time and optimal solution for the viewed paths. Moreover, it is shown that the computational effort is increasing slightly in the case of smooth joint jerk trajectory planning.

The drastic decrease in computation time indicates the presented approach as good candidate for online applications. By using methods similar to them used in model predictive control, the computation time could be further decreased. Using the presented approach in an online application framework will be addressed in future work.

\subsubsection*{Acknowledgement}
This research was funded in whole, or in part, by the Austrian Science Fund (FWF) [I 4452-N].


\bibliographystyle{splncs04}
\bibliography{RAAD_2023_lib}

\end{document}